\documentclass[runningheads]{llncs}
\usepackage{graphicx}
\usepackage{comment}
\usepackage{amsmath,amssymb} 
\usepackage{color}

\usepackage{booktabs}
\graphicspath{ {images/} }

\begin{document}
\pagestyle{headings}
\mainmatter
\def\ECCVSubNumber{1258}  

\title{GeoConv: Geodesic Guided Convolution for \\
Facial Action Unit Recognition} 

\titlerunning{GeoConv: Geodesic Guided Convolution for Facial Action Unit Recognition}
%
\author{Yuedong Chen\inst{1}\and
Guoxian Song\inst{1} \and
Zhiwen Shao\inst{2} \and \\
Jianfei Cai\inst{3} \and
Tat-Jen Cham\inst{1} \and
Jianmin Zheng\inst{1}}
\authorrunning{Yuedong Chen et al.}
%
\institute{Nanyang Technological University, Singapore \and
Shanghai Jiao Tong University, China \and
Monash University, Australia \\
\email{\{donald.chen,astjcham,asjmzheng\}@ntu.edu.sg, guoxian001@e.ntu.edu.sg \\
shaozhiwen@sjtu.edu.cn, jianfei.cai@monash.edu}}
\maketitle

\begin{abstract}
Automatic facial action unit (AU) recognition has attracted great attention but still remains a challenging task, as subtle changes of local facial muscles are difficult to thoroughly capture. Most existing AU recognition approaches leverage geometry information in a straightforward 2D or 3D manner, which either ignore 3D manifold information or suffer from high computational costs. In this paper, we propose a novel geodesic guided convolution (GeoConv) for AU recognition by embedding 3D manifold information into 2D convolutions. Specifically, the kernel of GeoConv is weighted by our introduced geodesic weights, which are negatively correlated to geodesic distances 
on a coarsely reconstructed 3D face model. Moreover, based on GeoConv, we further develop an end-to-end trainable framework named GeoCNN for AU recognition. 
Extensive experiments on BP4D and DISFA benchmarks show that our approach significantly outperforms the state-of-the-art AU recognition methods.

\end{abstract}

\section{Introduction}
Facial action unit (AU) classifies subtle movements of facial muscles, which is defined by the Facial Action Coding System (FACS)~\cite{friesen1978facial}. Unlike facial expressions, which categorize faces into a few general emotions, AUs capture small differences and subtle changes in facial appearance. AU recognition is a more challenging problem, and plays a critical role in human expression analysis and generation. 

Automatic AU recognition has attracted great attention in the research community.
Traditional approaches~\cite{valstar2006fully,bayramoglu2013cs} resort to hand-crafted features for AU recognition, which have a limited capacity in capturing subtle muscle actions. 
Recently there has been a rapid surge in
deep learning based methods applied to this problem. In particular, some methods extract discriminative features from correlated AU regions~\cite{li2018eac,corneanu2018deep,li2019semantic}, while others exploit 2D geometric information formed by facial landmarks~\cite{shao2018deep,niu2019local}. Although focusing on landmark-based 2D geometry is intuitive and has led to good progress, it does not capture 3D geometric information 
and may fail to discriminate fine-grained AUs. 

Recently, a few works exploit 3D structure to facilitate AU recognition. Liu et al.~\cite{liu2019conditional} utilized the 3D morphable model (3DMM)~\cite{blanz1999morphable} to synthesize various expressions for augmenting training data, but data augmentation does not directly incorporate 3D information in the learning process. Reale et al.~\cite{reale2019facial} extracted higher-resolution features from 3D point clouds, but 3D point clouds are not typically available and require high computational cost.

On the other hand, with advances in 3D face reconstruction from a single RGB image (e.g.~\cite{guo2018cnn}), it is possible to perform AU recognition directly on the reconstructed 3D face model by using 3D convolution operations. However, such an approach faces two major obstacles. First, high quality fine-grained 3D face reconstruction from a single RGB image is still challenging. Second, 3D convolution operations are typically time-consuming and memory-hungry with limited resolutions.

\begin{figure}[t!]
    \centering
    \includegraphics[width=0.7\textwidth]{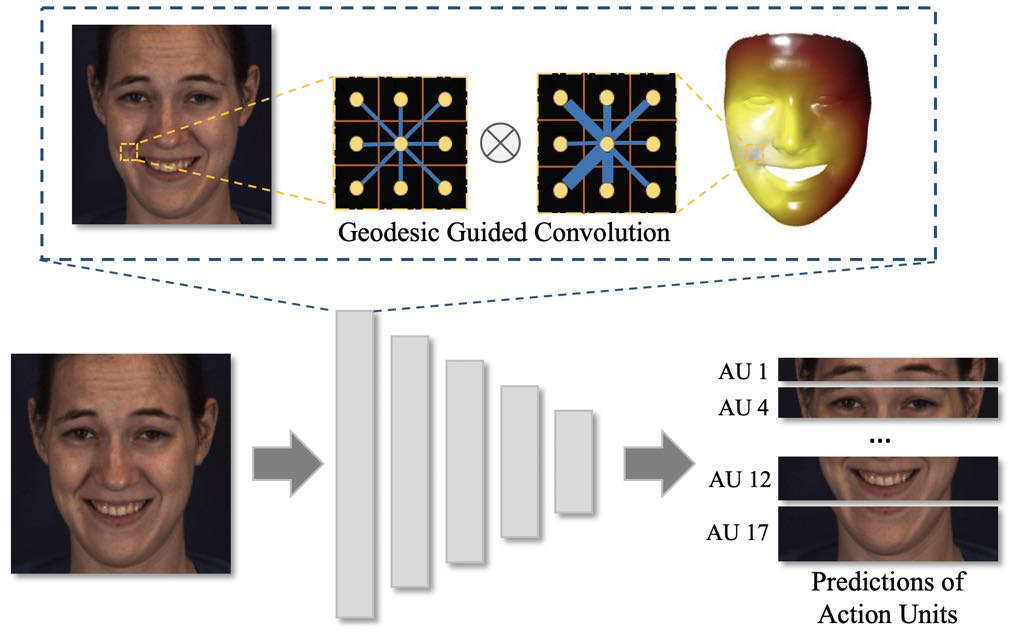}
    \caption{The overview of GeoConv for AU recognition. GeoConv embeds information from a coarse 3D face model (in the upper-right corner) into convolution operations by weighting kernels with the introduced geodesic weights. 
    GeoConv can replace its regular counterpart and be integrated into any existing CNN models for face analysis tasks such as AU recognition.}
    \label{fig:geoconv_overview}
\end{figure}

Motivated by the above considerations, in this research we propose to exploit 3D manifold information from coarsely reconstructed 3D facial surfaces while preserving the simplicity and efficiency of 2D operations. Specifically we propose a new convolution operation, called \textbf{Geodesic Guided Convolution (GeoConv)}, to achieve this goal. As shown in Fig.~\ref{fig:geoconv_overview}, in GeoConv the convolutional kernel is weighted by geodesic distances on the 3D facial surface. Our basic idea is that for a pixel/element, the contribution of a neighbor in a convolution should be negatively correlated to the geodesic distance. In this way, GeoConv can guide the training of a deep neural network with the awareness of the 3D facial manifold. The computation of geodesic distances only requires a coarse 3D model, which can be recovered from a 2D image efficiently~\cite{guo2018cnn}. Moreover, we introduce a hierarchy strategy to enable GeoConv to be integrated into any deep convolutional neural network (CNN). 

The contributions of this paper are summarized as follows.
\begin{itemize}
  \item We propose a novel geodesic guided convolution, which can be applied to fine-grained face analysis tasks such as AU recognition. 
  \item We propose an end-to-end trainable framework (named GeoCNN) based on GeoConv with the proposed hierarchy strategy for AU recognition. To our knowledge, this is the first work of integrating 3D information into convolution operations for AU recognition.
  \item Extensive experiments verify the effectiveness of our proposed GeoCNN and show that our method soundly outperforms the state-of-the-art AU recognition methods on BP4D and DISFA benchmarks.
\end{itemize}

\section{Related Work}
Automatic AU recognition has attracted much attention in the past decade. In this section, we mainly introduce works that are closely related to our approach. Besides, we review some works using specially designed convolutions for certain computer vision tasks. 

\noindent\textbf{2D Geometry based AU Recognition.} Since AUs are defined as the subtle movements of certain local regions on a human face, taking geometry information into consideration is an effective way to improve the performance of AU recognition. A typical approach is to localize different AUs based on their geometric location, either in image space~\cite{corneanu2018deep} or in feature space~\cite{li2018eac,li2019semantic}. Shao et al.~\cite{shao2018deep} proposed to jointly perform AU recognition and face alignment so as to use the precise AU locations provided by landmarks, and Shao et al.~\cite{shao2019facial} further captured the AU-related local features through a spatial attention mechanism. Niu et al.~\cite{niu2019local} improved the AU recognition by leveraging facial shape information extracted from landmarks. Although 2D geometry has been widely exploited, it has a limited capacity of capturing local geometry details for AU detection.       

\noindent\textbf{3D Geometry based AU Recognition.} 3D structure can model more fine-grained facial appearance changes than a 2D image, it has been proved to be useful in face analysis task and has attracted increasing attention in recent years. Bayramoglu et al.~\cite{bayramoglu2013cs} applied the LBP methodology to the 3D face points, and constructed a new operator to encode the geometric properties, which demonstrates the improvement of overall AU detection performance. Tulyakov et al.~\cite{tulyakov2015facecept3d} introduced FaceCept3D to build a 3D template from RGB-D data, which is then used to extract features for AU recognition. Reale et al.~\cite{reale2019facial} utilized unordered 3D point clouds and proposed a new architecture based on PointNet~\cite{qi2017pointnet} to directly consume the 3D data. More recently, Liu et al.~\cite{liu2019conditional} built an AU synthesis pipeline based on 3DMM~\cite{blanz1999morphable} to augment the current AU dataset, which is claimed to be effective in improving AU intensity estimation. The above methods have shown the effectiveness of applying 3D data to AU related task, but none of them explore incorporating 3D information into 2D convolution, which could be a more efficient solution because 2D convolution is more lightweight and well-developed than 3D based operations.

\noindent\textbf{Geometric Transformations in CNNs.} Because of the fixed structure of the convolutional kernel, the regular convolution has difficulty in capturing different local details. To tackle the limitation, Yu et al.~\cite{yu2015multi} introduced the dilation convolution by generalizing the regular convolution with a dilation factor, which can enlarge the receptive field and achieve better performance especially in dense prediction tasks. Dai et al.~\cite{dai2017deformable} proposed the deformable convolution to learn additional offsets, which can augment the spatial sampling location and perform well in object detection. Wang et al.~\cite{wang2018depth} presented depth-aware convolution to embed in convolution the information of depth similarity between pixels, which demonstrates superiority in RGB-D semantic segmentation. Although these methods revealed that augmenting regular convolution with geometric related design can boost the performance on several computer vision tasks, they do not incorporate 3D surface information into the convolution operations. Our work, by introducing GeoConv, is the first that incorporates 3D geometry information from a coarsely reconstructed 3D face model, and we further develop GeoCNN for AU recognition.


\section{Methodology}

In this section, we first introduce the background on 3D face modeling, and then describe our proposed GeoConv. Finally, we present our framework for AU recognition and we focus on the application of GeoConv.

\subsection{3D Face Modeling}

\textbf{3D Parametric Face Model.} In this work, we use the 3D Morphable Model (3DMM)~\cite{blanz1999morphable} as the 3D face representation. 3DMM encodes 3D face geometry 
into a low-dimensional subspace, and covers facial identity variations and expression variations by adding delta blendshapes to its shape model. Specifically, 3DMM describes 3D face geometry $p$ with Principal Component Analysis (PCA) as
\begin{align}
p(w^{id}, w^{exp}) &= \bar{p} + E^{id}w^{id} + E^{exp} w^{exp}, \label{eq:3dmm_1}
\end{align}
where $\bar{p}$ represents the average 3D face, $E^{id}$ and $E^{exp}$ respectively denote the principal axes of the identity space 
and the expression space, 
and $w^{id}$ and $w^{exp}$ are the corresponding coefficients. In our experiments, $w^{id}$ has 100 dimensions with the bases $E^{id}$ from the Basel Face Model (BFM)~\cite{paysan20093d}, while $w^{exp}$ has 79 dimensions with the bases $E^{exp}$ from FaceWarehouse~\cite{cao2014facewarehouse}.
 
\noindent\textbf{Face Inverse Rendering.} The rendering process is to project a 3D model onto a 2D image plane, while face inverse rendering is the inverse process of face image generation, i.e.\ given a 2D face image, a 3D face model, albedo, lighting condition, pose and projection parameters are expected to be estimated simultaneously. Directly obtaining all the above components with only one single input image is an ill-posed problem. Therefore, a typical way is to utilize the 3DMM parametric face model as a prior. Our work 
estimates the 3DMM parameters $w^{id}$ and $w^{exp}$ from a 2D face image so as to recover the 3D face model. 


\subsection{Geodesic Guided Convolution}

Convolution is one of the most basic operations in CNNs. In the 2D image space, convolution on a specific pixel location $x_0$ is conducted via a  weighted sum over all pixels within a local grid. Specifically, the convolution output $\mathcal{C}(x_0)$ is computed as
\begin{equation}
    \label{eq:std_conv}
    \mathcal{C}(x_0) = \sum_{x_i \in \mathcal{R}(x_0)} w(x_i)\mathcal{I}(x_i), 
\end{equation}
where $\mathcal{R}(x_0)$ is the local grid centered around $x_0$, and $w(x_i)$ and $\mathcal{I}(x_i)$ represent the kernel weight and the pixel value at location $x_i$, respectively.

\begin{figure}[t!]
    \centering
    \includegraphics[width=0.7\textwidth]{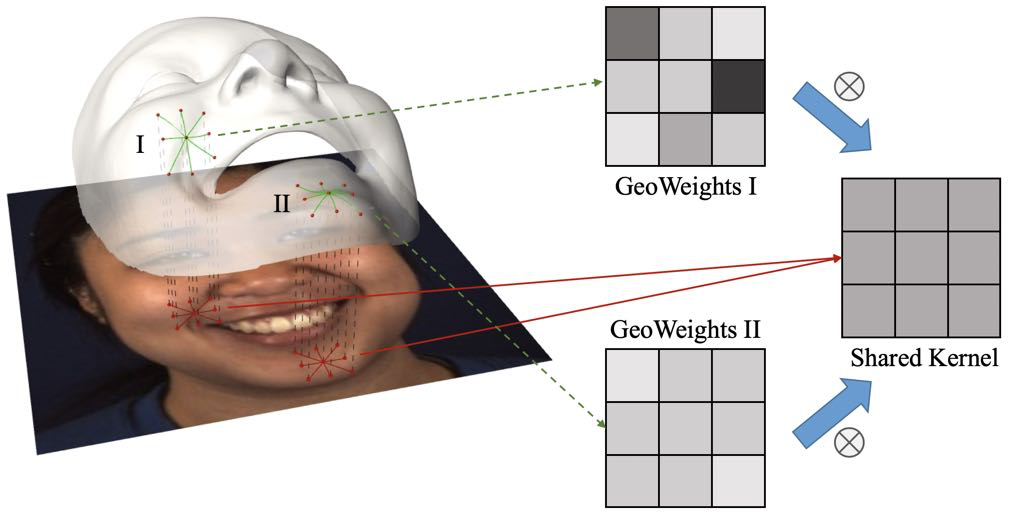}
    \caption{Illustration of information propagation in GeoConv.
    Kernel weights are  shared across the whole image plane in regular convolution, which may fail to capture fine-grained feature for different local regions. The proposed GeoConv addresses this issue by assigning different geodesic weights to different regions, see region I and region II in the 3D face model. 
    }
    \label{fig:geoconv}
\end{figure}

Typically when applying the convolution operation to an RGB image, the kernel weights of the convolution operation are shared across the whole image plane. Although it reduces trainable parameters and enforces translation invariance, treating different regions equally can be inappropriate, especially for tasks relying on fine-grained local details. Taking a face image as an example, as shown in Fig.~\ref{fig:geoconv}, a local grid around the lip corner and another one around the center of the chin would encounter the same kernel weights in a convolutional layer. It is clear that cells in the same locations of these two grids should contribute differently to their outputs, because the former is bumpy and the latter is flat. However, due to the lack of geometric information in 2D images and  shared kernel weights, the conventional convolution is not able to treat different local regions differently. For tasks that focus on the movements of local regions such as AU recognition, taking into consideration 3D geometric structure while conducting the 2D convolution operation may help extract more precise representation and boost performance.

Inspired by the above analysis, we therefore propose our new operation GeoConv that embeds 3D information into the 2D convolution by weighting the kernel (see Fig.~\ref{fig:geoconv}). In particular, geodesic distances between different vertices on the 3D face model obtained by face inverse rendering are employed to construct the weights. GeoConv is formulated as
\begin{equation}
    \mathcal{C}(x_0) = \sum_{x_i \in \mathcal{R}(x_0)} g(x_0, x_i) w(x_i)\mathcal{I}(x_i).  
\end{equation}
$g(x_0, x_i)$ denotes the geodesic weight, which is defined as
\begin{equation}
\label{eq:geo_weights}
g(x_0,x_i) = 
     \begin{cases}
      \ (N_{\mathcal{R}}-1)\times \sigma(-\frac{\mathcal{D}_{geo}(x_0, x_i)}{\mathcal{D}_{eu}(x_0, x_i)}), &\ \text{if } i \neq 0  \\
      \ e^{0}, &\ \text{if } i = 0 \\
     \end{cases},
\end{equation}
where $N_{\mathcal{R}}$ denotes the number of pixels in the local grid $\mathcal{R}$, $\sigma$ is the softmax function, $\mathcal{D}_{geo}(x_0, x_i)$ denotes the geodesic distance between the two points on the surface of the 3D face model corresponding to locations $x_0$ and $x_i$ on the image plane, and $\mathcal{D}_{eu}(x_0, x_i)$ represents the Euclidean distance between locations $x_0$ and $x_i$. 

Note that the geodesic weight is calculated based on the 3DMM model, which \textit{only} has correspondences to the input image.
For the $j$-th layer in a CNN, the geodesic weight of a specific location $x^{(j)}_i$ is obtained by projecting this location back to the input layer. Since the receptive field becomes larger as the network goes deeper, we obtain a corresponding local grid instead of a single position in the input layer for a higher layer location $x^{(j)}_i$. To resolve the ambiguity, the center of the local grid in the input layer is treated as the corresponding position for computing geodesic distances.
Although adding convolution layer can enlarge receptive fields, it does not change the center of receptive fields, nor the corresponding geodesic distance. But the geodesic distance between two neighbors will increase as the network uses pooling layer, so we further normalize the geodesic distance by the Euclidean distance between the centers of two neighboring receptive fields, denoted as $\mathcal{D}_{eu}(x_0, x_i)$ in Eq.~\eqref{eq:geo_weights}. In addition, $\mathcal{D}_{eu}(x_0, x_i)$ can also take into account the distance difference of diagonal neighbors, compared with horizontal or vertical neighbors.

The geodesic weight in Eq.~\eqref{eq:geo_weights} is essentially constructed based on the assumption that the further a cell on the 3D face surface is to the center, the less contribution it should have to the final convolution output. In the scenario when all the cells in the convolution grid are on a plane, their weights will all be normalized to 1, making the regular convolution a special case of our proposed GeoConv.

\subsection{Importance of GeoConv}
Recently, employing 3D geometry information to facilitate recognition tasks has attracted great attention, since it contains more detailed information. However, taking 3D data as input suffers from several major obstacles. Firstly, high-quality 3D data is not easy to capture directly. 
Recovering a 3D model from a 2D image is extremely challenging, especially for obtaining fine-grained 3D model suitable for recognition tasks. In addition, memory-efficient 3D representation such as 3D meshes  is not grid-based, for which well designed 2D CNNs cannot be applied directly. 

Our proposed GeoConv can be considered as a 2.5D operation. It leverages the 3D information while keeping 2D functionality, so as to embrace the advantages of both domains. In particular, GeoConv makes the 2D convolution aware of the 3D manifold surface. Since we only compute a rough geodesic distance on the surface, GeoConv only requires a coarse 3D model, which can be recovered from a 2D image efficiently. Besides, GeoConv does not include any additional trainable parameters, and can be flexibly integrated into the regular 2D convolution operation. Thus, it can be used in any existing CNN models. 

\subsection{GeoConv for AU Recognition}
\label{section:geocnn}
\begin{figure}[t!]
    \centering
    \includegraphics[width=0.995\textwidth]{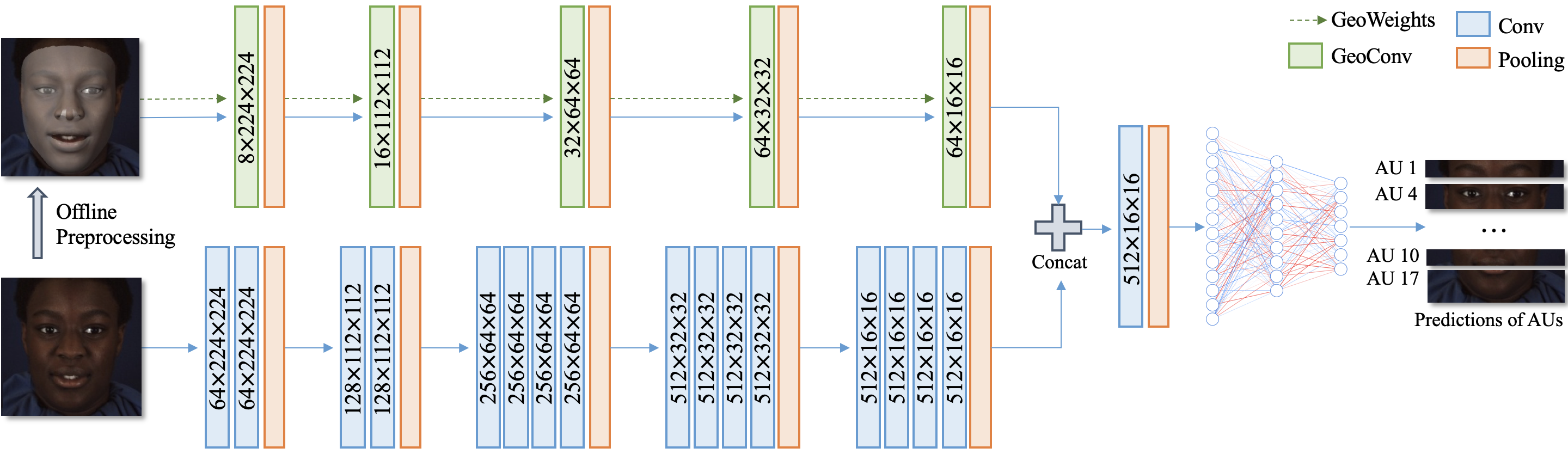}
    \caption{The architecture of the proposed GeoCNN. It follows a two-branch setting, in which the bottom branch is a VGG-19 network, and the top one is composed of the proposed GeoConv. Specific output feature size for each layer is illustrated as \textit{channels$\times$heights$\times$weights}, shown in each block. GeoCNN only takes a 2D image as input, and for a given face image, an efficient preprocess is used to estimate its corresponding 3DMM model parameters and compute geodesic distances on the 3D surface, which will be further leveraged to obtain geodesic weights for GeoConv. 
    }
    \label{fig:tb_model}
\end{figure}

Considering that there are close correlations between AUs and facial geometric structure, 
GeoConv is highly applicable for AU recognition.
We propose a novel deep learning based framework called GeoCNN for AU recognition. It follows a two-branch setting, as illustrated in Fig.~\ref{fig:tb_model}.
The branch at the bottom is a VGG-19~\cite{simonyan2014very} network, which is used to capture rich facial information. The branch at the top is composed of the proposed GeoConv layers, which are used to capture fine-grained AU information. The features extracted by two branches are then concatenated and integrated in the following layers. Finally, a fully-connected classifier is designed to predict the AU occurrence probabilities.

Specifically, the geodesic weights are obtained by the following preprocessing pipeline. Firstly, the input image is fed to a pretrained CoarseNet \cite{guo2018cnn} to estimate its 3DMM parameters (about 20ms), which are then directly used to construct the 3DMM model using Eq.~\eqref{eq:3dmm_1}. After that, the geodesic distances on the 3DMM model are computed using the heat-based algorithm~\cite{crane2017heat}, which takes about 260s in total for one input image. Finally, geodesic weights are calculated using Eq.~\eqref{eq:geo_weights}. This preprocessing can be further optimized. 

Considering the severe data imbalance issue~\cite{li2018eac,shao2018deep} in the training set, we propose to balance the loss function with weights calculated through analyzing the data distribution of the training set. The AU recognition loss is formulated as
\begin{equation}
\label{eq:loss_bce}
l_{au} = -\sum_{i=1}^{N_{au}}w_i[p_iy_i \log \hat{y_i} + (1-y_i) \log (1 - \hat{y_i} )], 
\end{equation}
where $N_{au}$ is the number of AUs, $y_i$ refers to the ground-truth probability for the $i$-th AU, $\hat{y_i}$ is the corresponding predicted occurrence probability, $w_i$ is the weight of the $i$-th AU, and $p_i$ represents the weight for the positive samples. Moreover, $w_i$ and $p_i$ are computed as follows: 
\begin{equation}
    w_i = \frac{n}{n_i^{pos}} \Big/ \textstyle \sum_{i=1}^{N_{au}} \frac{n}{n_i^{pos}}, \;\;\;
    p_i = \frac{n - n^{pos}_i}{n^{pos}_i}, 
\end{equation}
where $n$ and $n_i^{pos}$ denote the number of all samples and the number of positive samples of the $i$-th AU in the training set, respectively.

\section{Experiments}
\subsection{Datasets and Settings}

\noindent\textbf{Datasets.} The proposed approach is evaluated on two widely used spontaneous facial expression datasets, BP4D~\cite{zhang2013high} and DISFA~\cite{mavadati2013disfa}. 

\textbf{BP4D} includes 41 young adult subjects with 23 women and 18 men, each of whom is associated with 8 video-recorded tasks. Around 140,000 frames are manually annotated with binary FACS AU labels. As a general approach~\cite{zhao2016deep,shao2018deep,li2019semantic}, 12 AUs are measured with the subject-independent 3-fold cross validation, where two folds are used for training with the remaining one for testing in turns.

\textbf{DISFA} consists of videos of 27 adult participants of different ethnicities, with 12 females and 15 males. All video frames are manually scored with the intensity of AU, ranging from 0 to 5. Following the settings of~\cite{zhao2018learning,li2019semantic}, we set AUs with intensity equal or larger than 2 as positive, and others as negative. We refer to~\cite{zhao2018learning,li2019semantic} to evaluate 8 AUs with subject-exclusive 3-fold cross validation,  similar to BP4D.

\noindent\textbf{Implementation Details.} For each input face image, shape-preserving transformation is applied to it before being fed to the network. Specifically, after estimating the 3DMM model of an image, we extract the 2D facial region by rasterization. Images are then aligned according to the center positions of their corresponding facial landmarks, and finally cropped into size $224\times224$.
During training, two kinds of online data augmentation are considered. Hue, saturation and brightness of an input image are scaled with coefficients uniformly sampled from [0.6, 1.4]. Noise jitter is also applied through adding PCA noise with a coefficient drawn from a Gaussian distribution $\mathcal{N}(0, 0.1)$. 

The VGG-19 branch is initialized with ImageNet~\cite{deng2009imagenet} pretrained weights, while the remaining trainable parameters are randomly initialized. The model is trained with Stochastic Gradient Descent (SGD) optimizer, where momentum factor and L2 penalty are set to 0.9 and 0.0005, respectively, and Nesterov momentum~\cite{sutskever2013importance} is enabled. The initial learning rates for the VGG-19 branch and other trainable parameters are set to 0.001 and 0.0005, respectively. We trained the whole model for 14 epochs, during which learning rates are decayed using a cosine annealing~\cite{loshchilov2016sgdr} strategy. To maintain rigor and fairness of comparison in the experiments we conducted, all settings such as data augmentation, learning rate, etc.\ remain the same as the full model, except for changes that are explicitly stated.
We implemented the proposed GeoCNN model using PyTorch~\cite{paszke2019pytorch}, and GeoConv operator using PyTorch extension with CUDA acceleration. 

\noindent\textbf{Evaluation Metric.} Our method is evaluated with the frame-based F1-score (F1-frame, \%), which has been widely employed in AU recognition and other binary classification tasks. It is computed as the harmonic mean of the precision and the recall. We report the final results averaged over 3-fold tests in all the following experiments. For the sake of simplicity, we omit \% in the following sections regarding F1-frame.

\subsection{Comparisons with State-of-the-Art Methods}

We validate our proposed framework by comparing it with the state-of-the-art AU recognition approaches under the same subject-exclusive three-fold cross validation settings. Specifically, these approaches are LSVM~\cite{fan2008liblinear}, JPML~\cite{zhao2015joint}, DRML~\cite{zhao2016deep}, EAC-Net~\cite{li2018eac}, DSIN~\cite{corneanu2018deep}, CMS~\cite{sankaran2019representation}, JAA-Net~\cite{shao2018deep}, SRERL~\cite{li2019semantic}, LP-Net~\cite{niu2019local} and ARL~\cite{shao2019facial}. Note that we only consider image-based methods and ignore sequence-based methods due to our experiment settings. Furthermore, we noticed that a threshold tuning strategy is leveraged by DSIN. Since this strategy is not used by other methods, we chose to report all results based on the standard threshold for fair comparison. For LSVM and JPML, we used their results as reported in \cite{shao2018deep,li2019semantic}. The results of other methods were directly obtained from their corresponding papers.

\begin{table}[t!]
\centering
\caption{F1-frame results of 12 AUs
for the proposed GeoCNN and the state-of-the-art approaches
on BP4D dataset. EAC, JAA and LP refer to EAC-Net, JAA-Net and LP-Net, respectively. The best and the second best are indicated using bold fonts and brackets, respectively.}
\begin{tabular}{@{}lccccccccccc@{}}
\toprule
Method & LSVM & JPML          & DRML & EAC  & DSIN          & JAA           & CMS           & SRERL         & LP         & ARL           & \textbf{GeoCNN} \\ \midrule \midrule
AU1    & 23.2 & 32.6          & 36.4 & 39.0 & \textbf{51.7} & 47.2          & {[}49.1{]}    & 46.9          & 43.4       & 45.8          & 48.4            \\
AU2    & 22.8 & 25.6          & 41.8 & 35.2 & 40.4          & 44.0          & 44.1          & \textbf{45.3} & 38.0       & 39.8          & {[}44.2{]}      \\
AU4    & 23.1 & 37.4          & 43.0 & 48.6 & {[}56.0{]}    & 54.9          & 50.3          & 55.6          & 54.2       & 55.1          & \textbf{59.9}   \\
AU6    & 27.2 & 42.3          & 55.0 & 76.1 & 76.1          & 77.5          & \textbf{79.2} & 77.1          & 77.1       & 75.7          & {[}78.4{]}      \\
AU7    & 47.1 & 50.5          & 67.0 & 72.9 & 73.5          & 74.6          & 74.7          & \textbf{78.4} & 76.7       & {[}77.2{]}    & 75.6            \\
AU10   & 77.2 & 72.2          & 66.3 & 81.9 & 79.9          & \textbf{84.0} & 80.9          & 83.5          & {[}83.8{]} & 82.3          & 83.6            \\
AU12   & 63.7 & 74.1          & 65.8 & 86.2 & 85.4          & 86.9          & \textbf{88.3} & {[}87.6{]}    & 87.2       & 86.6          & 86.7            \\
AU14   & 64.3 & \textbf{65.7} & 54.1 & 58.8 & 62.7          & 61.9          & 63.9          & 63.9          & 63.3       & 58.8          & {[}65.0{]}      \\
AU15   & 18.4 & 38.1          & 33.2 & 37.5 & 37.3          & 43.6          & 44.4          & {[}52.2{]}    & 45.3       & 47.6          & \textbf{53.0}   \\
AU17   & 33.0 & 40.0          & 48.0 & 59.1 & 62.9          & 60.3          & 60.3          & {[}63.9{]}    & 60.5       & 62.1          & \textbf{64.7}   \\
AU23   & 19.4 & 30.4          & 31.7 & 35.9 & 38.6          & 42.7          & 41.4          & 47.1          & {[}48.1{]} & 47.4          & \textbf{49.5}   \\
AU24   & 20.7 & 42.3          & 30.0 & 35.8 & 41.6          & 41.9          & 51.2          & 53.3          & {[}54.2{]} & \textbf{55.4} & 54.1            \\ \midrule
Avg.   & 35.3 & 45.9          & 48.3 & 55.9 & 58.9          & 60.0          & 60.6          & {[}62.9{]}    & 61.0       & 61.1          & \textbf{63.6}   \\ \bottomrule
\end{tabular}
\label{tab:bp4d_sota}
\end{table}

\begin{table}[h!]
\centering
\caption{F1-frame results of 8 AUs 
for the proposed GeoCNN and existing state-of-the-art approaches 
on the DISFA dataset. The best and the second best are indicated using bold fonts and brackets, respectively.}
\begin{tabular}{@{}lcccccccccc@{}}
\toprule
Method & LSVM & DRML & EAC       & DSIN       & JAA-Net & CMS           & SRERL      & LP-Net        & ARL           & \textbf{GeoCNN} \\ \midrule \midrule
AU1    & 10.8 & 17.3 & 41.5          & 42.4       & 43.7    & 40.2          & {[}45.7{]} & 29.9          & 43.9          & \textbf{65.5}   \\
AU2    & 10.0 & 17.7 & 26.4          & 39.0       & 46.2    & 44.3          & {[}47.8{]} & 24.7          & 42.1          & \textbf{65.8}   \\
AU4    & 21.8 & 37.4 & 66.4          & {[}68.4{]} & 56.0    & 53.2          & 59.6       & \textbf{72.7} & 63.6          & 67.2            \\
AU6    & 15.7 & 29.0 & {[}50.7{]}    & 28.6       & 41.4    & \textbf{57.1} & 47.1       & 46.8          & 41.8          & 48.6            \\
AU9    & 11.5 & 10.7 & \textbf{80.5} & 46.8       & 44.7    & 50.3          & 45.6       & 49.6          & 40.0          & {[}51.4{]}      \\
AU12   & 70.4 & 37.7 & \textbf{89.3} & 70.8       & 69.6    & 73.5          & 73.5       & 72.9          & {[}76.2{]}    & 72.6            \\
AU25   & 12.0 & 38.5 & 88.9          & 90.4       & 88.3    & 81.1          & 84.3       & {[}93.8{]}    & \textbf{95.2} & 80.9            \\
AU26   & 22.1 & 20.1 & 15.6          & 42.2       & 58.4    & 59.7          & 43.6       & {[}65.0{]}    & \textbf{66.8} & 44.9            \\ \midrule
Avg.   & 21.8 & 26.7 & 48.5          & 53.6       & 56.0    & 57.4          & 55.9       & 56.9          & {[}58.7{]}    & \textbf{62.1}   \\ \bottomrule
\end{tabular}
\label{tab:disfa_sota}
\end{table}

Table~\ref{tab:bp4d_sota} shows the quantitative results, averaged over three runs, of different approaches on the BP4D dataset. Overall, our GeoCNN model achieved the best average F1-frame result, with the best or second best scores on most of the tested AUs. Traditional approaches, namely LSVM and JPML, leverage hand-crafted features which are less representative, and result in much worse performance. Compared with the latest deep learning based methods, our model in general outperformed all, achieving the best in 4 out of 12 AUs. This is mainly because that our method can capture subtle facial muscle changes and extract fine-grained features with the help of 3D information.

Table~\ref{tab:disfa_sota} shows the F1-frame results of different methods on the DISFA dataset. It can be seen that our model outperformed all state-of-the-art approaches by a large margin, with an increase of 3.4\% over the second best approach ARL in terms of the average F1-frame. This large gain is mainly contributed by the large improvement on AU1 (Inner Brow Raiser) and AU2 (Outer Brow Raiser). Specifically, the raising of brow might be too subtle to be seen on the 2D plane, while it is noticeable on the 3D surface due to the change of depth around the area of the eyes. We also noticed that our model did not perform well on AU25 (Lips Part) and AU26 (Jaw Drop). This is mainly because the two AUs only exist when the mouth is open, where the disconnect between the two lip regions in 3DMM leads to a very large geodesic distance, causing a dramatically different geodesic weight pattern from other AU regions. 

\subsection{Ablation Study}
In this section, we evaluate individual components in the proposed GeoCNN.
As shown in Fig.~\ref{fig:tb_model}, GeoCNN has 5 GeoConv layers. For better exploration and illustration, in the following parts, we name a few GeoCNN variants as $\mathcal{G}_{(iiiii)}$, where $i$ is set to 1 if the corresponding layer is GeoConv and 0 if it is regular convolution, e.g. GeoCNN is equivalent to $\mathcal{G}_{(11111)}$. Table~\ref{tab:bp4d_ablation} shows the quantitative comparison results of different configurations. 

\begin{table}[th]
\centering
\caption{F1-frame results of 12 AUs on BP4D dataset for ablation study. For GeoCNN variants labeled as $\mathcal{G}_{(iiiii)}$, $i$ is set to 1 if the corresponding layer is GeoConv and 0 if it is regular convolution. GeoCNN is the same as $\mathcal{G}_{(11111)}$. BW denotes using the balanced weights in the loss function. HC denotes using hierarchy compensation in geodesic weight. The best results are indicated in bold fonts.}
\begin{tabular}{@{}lccccccccccccc@{}}
\toprule
Method                  & AU1           & AU2           & AU4           & AU6           & AU7           & AU10          & AU12          & AU14          & AU15          & AU17          & AU23          & AU24          & Avg.          \\ \midrule \midrule
VGG-19                  & 49.1          & 39.3          & 54.2          & 76.4          & 73.6          & 80.9          & 85.8          & 55.3          & 40.4          & 58.8          & 33.2          & 45.0          & 57.7          \\
$\mathcal{G}_{(00000)}$ & 45.5          & 43.8          & 51.6          & 77.7          & 74.6          & \textbf{84.1} & \textbf{88.1} & 60.8          & 36.3          & 58.5          & 35.2          & 43.5          & 58.3          \\ \midrule
$\mathcal{G}_{(10000)}$ & 49.3          & 41.8          & 55.7          & 78.3          & 75.8          & 81.6          & 86.6          & 59.8          & 50.2          & 62.9          & 47.2          & 43.9          & 61.1          \\
$\mathcal{G}_{(00001)}$ & 45.8          & 38.8          & 56.6          & 77.9          & 75.7          & 80.9          & 86.9          & 61.8          & 43.7          & 60.5          & 47.8          & 42.8          & 59.9          \\
$\mathcal{G}_{(11100)}$ & \textbf{53.2} & 41.7          & 57.9          & 77.7          & 76.3          & 83.1          & 87.7          & \textbf{65.1} & 49.2          & 61.9          & 44.0          & 48.5          & 62.2          \\
$\mathcal{G}_{(00111)}$ & 49.2          & 41.1          & 59.4          & \textbf{78.9} & 75.2          & 80.8          & 86.8          & 60.5          & 47.8          & 62.1          & 46.6          & 50.2          & 61.6          \\ \midrule
w/o HC                  & 48.9          & 39.6          & 54.8          & 76.5          & \textbf{77.2} & 81.1          & 86.0          & 61.5          & 50.1          & 62.3          & 42.3          & 46.9          & 60.6          \\ \midrule
w/o BW                  & 47.0          & 39.9          & 55.4          & 78.8          & 76.3          & 82.2          & 86.8          & 63.9          & 49.4          & 62.7          & 47.6          & 50.6          & 61.7          \\ \midrule
GeoCNN                  & 48.4          & \textbf{44.2} & \textbf{59.9} & 78.4          & 75.6          & 83.6          & 86.7          & 65.0          & \textbf{53.0} & \textbf{64.7} & \textbf{49.5} & \textbf{54.1} & \textbf{63.6} \\ 
\bottomrule
\end{tabular}
\label{tab:bp4d_ablation}
\end{table}

\noindent\textbf{Baselines.} We consider two baselines for our model. One is using only the VGG-19 branch, and the other is using the full two-branch model but replacing all GeoConv with its counterpart using regular convolution, denoted as $\mathcal{G}_{(00000)}$. As shown in Table~\ref{tab:bp4d_ablation}, compared to the single branch VGG-19 model, model $\mathcal{G}_{(00000)}$ led to a minor improvement with a 0.6\% increase in terms of average F1-frame. As illustrated in Fig.~\ref{fig:tb_model}, the geodesic branch is shallower than the VGG-19 branch. Fusing features extracted by two different branches can help improve the performance, thus leading to a slightly better classification result. In contrast, GeoCNN outperformed both VGG-19 and $\mathcal{G}_{(00000)}$ significantly, improving by 5.9\% and 5.3\%, respectively, which justify the effectiveness of GeoConv. 

\noindent\textbf{GeoConv Locations.} To explore the locations to best add GeoConv, we set up four comparative experiments, including model 
$\mathcal{G}_{(10000)}$, $\mathcal{G}_{(00001)}$, $\mathcal{G}_{(11100)}$ and $\mathcal{G}_{(00111)}$. As illustrated in Table~\ref{tab:bp4d_ablation}, compared to $\mathcal{G}_{(10000)}$, $\mathcal{G}_{(11100)}$ adds more GeoConv, which helps propagate the geometric relationships and improve AU recognition. Similar pattern can also be found by comparing $\mathcal{G}_{(00111)}$ to $\mathcal{G}_{(00001)}$. Besides, $\mathcal{G}_{(00001)}$ gave worse results than $\mathcal{G}_{(10000)}$ and  $\mathcal{G}_{(00111)}$ gave worse results than $\mathcal{G}_{(11100)}$, 
which suggest that GeoConv performs better in lower layers instead of higher layers. This is understandable since GeoConv is highly correlated to structural information, which may be reduced in the previous pooling layers. By adding five layers of GeoConv throughout the additional branch, GeoCNN achieve the best performance. 

\noindent\textbf{Hierarchy Compensation.} Since geodesic distance will increase due to network hierarchy, we introduce Euclidean distance $\mathcal{D}_{eu}(x_0, x_i)$ to compensate for it, as shown in Eq.~\eqref{eq:geo_weights}. To validate the effectiveness, our models were compared with and without hierarchy compensation, where the latter is referred to as ``w/o HC''. Specifically, we replaced $\mathcal{D}_{eu}(x_0, x_i)$ with 1.0 in w/o HC. Comparing w/o HC with GeoCNN, it can be seen that introducing Euclidean distance to normalize geodesic distance has an important impact, leading to an improvement of 3.0\% in terms of average F1-frame, with such improvement also seen in the performance of almost every single AU. The results justify that the introduced hierarchy compensation helps stabilize the information propagation of GeoConv throughout the network. 

\noindent\textbf{Balanced Weights.} To address AU data imbalance, we introduced the balanced weights $w_i$ and $p_i$ in the loss function in Eq.~\eqref{eq:loss_bce}. To evaluate the effectiveness, we compared our models with and without the balanced weights, where the latter was denoted by ``w/o BW''. Comparing w/o BW with GeoCNN, we can see that the balanced loss led to a better performance, since the balanced weights help tackle both the inter-class and intra-class AU imbalance problems.

\subsection{Analysis on GeoCNN Architecture}
We also conducted experiments to validate whether the proposed two-branch architecture, including a shallower GeoConv branch and a deeper backbone, is a good choice. Specifically, we considered a few architecture alternatives, either based on a single-branch setting or a two-branch setting. 
Table~\ref{tab:model_design} gives the results under different network architectures, all reported on the BP4D dataset.

\begin{table}[th]
\centering
\caption{F1-frame results of 12 AUs  on BP4D dataset for different network architectures. The suffix ``s'' denotes training from scratch. The best are indicated using bold fonts.}
\begin{tabular}{@{}lccccccccccccc@{}}
\toprule
Method  & AU1           & AU2           & AU4           & AU6           & AU7           & AU10          & AU12          & AU14          & AU15          & AU17          & AU23          & AU24          & Avg.          \\ \midrule \midrule
VGG-19  & 49.1          & 39.3          & 54.2          & 76.4          & 73.6          & 80.9          & 85.8          & 55.3          & 40.4          & 58.8          & 33.2          & 45.0          & 57.7          \\
Geo-19  & 40.5          & 36.6          & 43.2          & 75.8          & 72.2          & 79.2          & 85.4          & 59.8          & 41.1          & 58.3          & 37.3          & 47.0          & 56.4          \\
VGG-19s & 30.7          & 23.2          & 37.9          & 77.4          & 67.4          & 81.7          & 83.1          & 56.2          & 26.7          & 51.2          & 17.6          & 38.2          & 49.3          \\
Geo-19s & 37.0          & 34.8          & 39.4          & 75.1          & 71.8          & 80.4          & 83.4          & 61.1          & 37.6          & 55.7          & 29.8          & 41.7          & 54.0          \\ \midrule
DeepTB  & \textbf{52.8} & \textbf{46.6} & 57.7          & 77.9          & \textbf{76.6} & 82.3          & 86.0          & 63.2          & 48.5          & 62.4          & 49.2          & 53.2          & 63.0          \\
GeoCNN  & 48.4          & 44.2          & \textbf{59.9} & \textbf{78.4} & 75.6          & \textbf{83.6} & \textbf{86.7} & \textbf{65.0} & \textbf{53.0} & \textbf{64.7} & \textbf{49.5} & \textbf{54.1} & \textbf{63.6} \\ \bottomrule
\end{tabular}
\label{tab:model_design}
\end{table}

\noindent\textbf{Single-Branch Setting.} As mentioned, GeoConv does not introduce any additional trainable parameters, and it can replace its regular counterpart with little extra effort. Given that VGG-19 is chosen as the backbone, rather than using the proposed two-branch architecture, a simple alternative is to integrate GeoConv into the VGG-19 backbone by replacing all regular convolutional layers in VGG-19 with GeoConv, which we denote by Geo-19. 

As shown in Table~\ref{tab:model_design}, Geo-19 performed a bit worse than VGG-19, with a 1.3\% gap. Our conjecture is that VGG-19 benefits much more from the pretrained model. Although VGG-19 and Geo-19 are both initialized with weights pretrained on ImageNet, the convolution kernels of GeoConv used in Geo-19 are further weighted by geodesic weights, which implicitly changes the pretrained kernels. That is, the pretrained model that is a local optimum for regular convolution, is unlikely to be a local optimum for GeoConv due to the introduced geodesic weights, therefore leading to worse results for Geo-19.

To verify the above conjecture, we further conducted experiments by training both VGG-19 and Geo-19 from scratch, named as VGG-19s and Geo-19s, respectively. Table~\ref{tab:model_design} shows that Geo-19s significantly outperformed VGG-19s with a large margin of 4.7\%. Besides, comparing Geo-19s with Geo-19, we can see that although Geo-19 outperformed Geo-19s, their results were comparable, while the difference between VGG-19 and VGG-19s was huge. In summary, ImageNet pretrained weights can benefit regular convolution more than GeoConv, and when compared fairly by training from scratch, GeoConv performed better. 

\noindent\textbf{Two-Branch Setting.} From the above experiments, we can see that, for AU recognition, it is important to keep the pretrained backbone unchanged. Therefore, our proposed GeoCNN uses the two-branch setting, with the pretrained VGG-19 backbone to extract semantic features, while using a shallow network branch with GeoConv. For the two-branch setting, instead of using a shallow GeoConv branch, another intuitive alternative would be setting both branches with the same deep network architecture, equivalent to directly combining Geo-19 and VGG-19, which is termed as Deeper Two Branch (DeepTB).

As illustrated in Table~\ref{tab:model_design}, with a deeper GeoConv branch, DeepTB did not outperform the proposed GeoCNN. Our conjecture is that the additional GeoConv branch is more to incorporate 3D information to capture fine-grained local features, which might not need a deeper network. Besides, due to fewer GeoConv layers, GeoCNN has fewer trainable parameters than DeepTB, resulting in less usage of memory and computation time. Therefore, considering better performance and higher efficiency, we decide to choose the proposed two-branch architecture for GeoCNN, with one relatively shallower GeoConv branch and one deeper backbone.

\subsection{GeoConv vs. RGBD}
Considering that the main idea of this work is to incorporate 3D geometry information into the 2D image grid based CNN process, another intuitive alternative is to directly convert the recovered 3DMM face model as an additional depth input, which can be directly fed into CNN without the need to change basic operations of the network. 


To study this alternative, we set up two RGBD-based baselines. One involves using VGG-19 but replacing its image input with the concatenation of an RGB image and its depth map, which is a typical way of using depth information. The other is using the two-branch setting similar to GeoCNN but feeding the additional branch with the depth map, which is a fairer comparison to GeoCNN. We refer to the former as RGBD, and the latter as RGBDTB (RGBD Two-branch). Both RGBD and RGBDTB use only regular convolution rather than GeoConv. Their F1-frame results are given in Table~\ref{tab:rgbd_compare}.

\begin{table}[th]
\centering
\caption{F1-frame results of 12 AUs on BP4D dataset for the comparisons with RGBD methods. D denotes depth and RGBDTB denotes RGBD Two-branch. The best are indicated in bold fonts.}
\begin{tabular}{@{}lccccccccccccc@{}}
\toprule
Method & AU1           & AU2           & AU4           & AU6           & AU7           & AU10          & AU12          & AU14          & AU15          & AU17          & AU23          & AU24          & Avg.          \\ \midrule \midrule
VGG-19 & \textbf{49.1} & 39.3          & 54.2          & 76.4          & 73.6          & 80.9          & 85.8          & 55.3          & 40.4          & 58.8          & 33.2          & 45.0          & 57.7          \\
RGBD   & 46.3          & 38.8          & 54.4          & 78.0          & 74.4          & 82.0          & 86.6          & 61.4          & 42.4          & 60.8          & 44.1          & 46.0          & 59.6          \\
RGBDTB & 47.0          & 41.9          & 58.9          & 75.8          & \textbf{76.0} & 83.0          & \textbf{87.4} & 61.4          & 46.6          & 60.7          & 47.1          & 45.7          & 61.0          \\
GeoCNN & 48.4          & \textbf{44.2} & \textbf{59.9} & \textbf{78.4} & 75.6          & \textbf{83.6} & 86.7          & \textbf{65.0} & \textbf{53.0} & \textbf{64.7} & \textbf{49.5} & \textbf{54.1} & \textbf{63.6} \\ \bottomrule
\end{tabular}
\label{tab:rgbd_compare}
\end{table}

It can be seen that both RGBD and RGBDTB performed better than the baseline VGG-19, improving by 1.9\% and 3.3\% in terms of average F1-frame, respectively. Similar to GeoConv, these outcomes show that geometry information can help improve the performance of AU recognition. 
On the other hand, our proposed GeoCNN outperformed both of the RGBD based methods by 4.0\% and 2.6\%, respectively. Moreover, GeoCNN achieved the best in 9 out of 12 AUs.
We believe that this is because it is more effective to integrate prior knowledge down to the kernel level, instead of directly feeding the depth map into the network and letting the network itself figure out how to make use of the 3D geometry information. 
The above results overwhelmingly demonstrate that the well-designed GeoConv is a better solution than the naive RGBD based approaches for AU recognition.

\section{Conclusion}
We have proposed a novel GeoConv operation by integrating 3D information into the 2D convolution without introducing additional trainable parameters. 
Furthermore, we have proposed an end-to-end trainable framework GeoCNN, which is beneficial for AU recognition.
Extensive experiments on two benchmark datasets of BP4D and DISFA have demonstrated that our framework significantly outperforms existing state-of-the-art AU recognition methods. 

There are a few interesting directions for future work. First, GeoConv is a general operation and it can be easily applied to other fine-grained facial analysis tasks. Second, the current model does not perform well for mouth related AUs when the mouth is open. Such extreme cases are worthy of further investigation.

%
%
\bibliographystyle{splncs04}
\bibliography{egbib}
\end{document}